\documentclass[final]{cvpr}

\usepackage{times}
\usepackage{epsfig}
\usepackage{graphicx}
\usepackage{amsmath}
\usepackage{amssymb}

\usepackage[pagebackref=true,breaklinks=true,colorlinks,bookmarks=false]{hyperref}

\def\cvprPaperID{699} 
\def\confYear{CVPR 2021}

\begin{document}

\title{AID: Pushing the Performance Boundary of Human Pose Estimation \\ with Information Dropping Augmentation}

\author{Junjie Huang,$^1$ Zheng Zhu,$^2$ Guan Huang,$^1$ Dalong Du$^1$ \\
$^1$XForwardAI Technology Co.,Ltd, Beijing, China\\
$^2$Tsinghua University, Beijing, China\\
{\tt\small \{junjie.huang, zhengzhu\}@ieee.org, \{guan.huang, dalong.du\}@xforwardai.com}
}

\maketitle

\begin{abstract}
Both appearance cue and constraint cue are vital for human pose estimation. However, there is a tendency in most existing works to overfitting the former and overlook the latter. In this paper, we propose \texttt{Augmentation by Information Dropping (AID)} to verify and tackle this dilemma. Alone with AID as a prerequisite for effectively exploiting its potential, we propose customized training schedules, which are designed by analyzing the pattern of loss and performance in training process from the perspective of information supplying. In experiments, as a model-agnostic approach, AID promotes various state-of-the-art methods in both bottom-up and top-down paradigms with different input sizes, frameworks, backbones, training and testing sets. On popular COCO human pose estimation \texttt{test} set, AID consistently boosts the performance of different configurations by around 0.6 AP in top-down paradigm and up to 1.5 AP in bottom-up paradigm. On more challenging CrowdPose dataset, the improvement is more than 1.5 AP. As AID successfully pushes the performance boundary of human pose estimation problem by considerable margin and sets a new state-of-the-art, we hope AID to be a regular configuration for training human pose estimators. The source code will be publicly available for further research.
\end{abstract}

\section{Introduction}
Human pose estimation serves many visual understanding tasks such as video surveillance \cite{li2019state} and action recognition \cite{carreira2017quo, zhu2019action, zhu2019convolutional}. In recent years, research community has witnessed a significant advance from single person \cite{PS,DPM,DeepPose,tompson2014joint,CPM,Hourglass,fppose} to multi-person pose estimation \cite{DeepCut,DeeperCut,OpenPose,G-RMI,CPN,HRNet,AssociativeEmbedding,Higher}, where the key engine consists of the network architecture evolution \cite{CPN,SBNet,HRNet,MSPN,RSN}, unbiased data processing \cite{DARK,UDP} and effective grouping strategies \cite{DeepCut,DeeperCut,OpenPose,AssociativeEmbedding}.

\begin{figure*}[t]
	\setlength{\abovecaptionskip}{0.cm}
    \begin{center}
        \includegraphics[width=0.85\hsize]{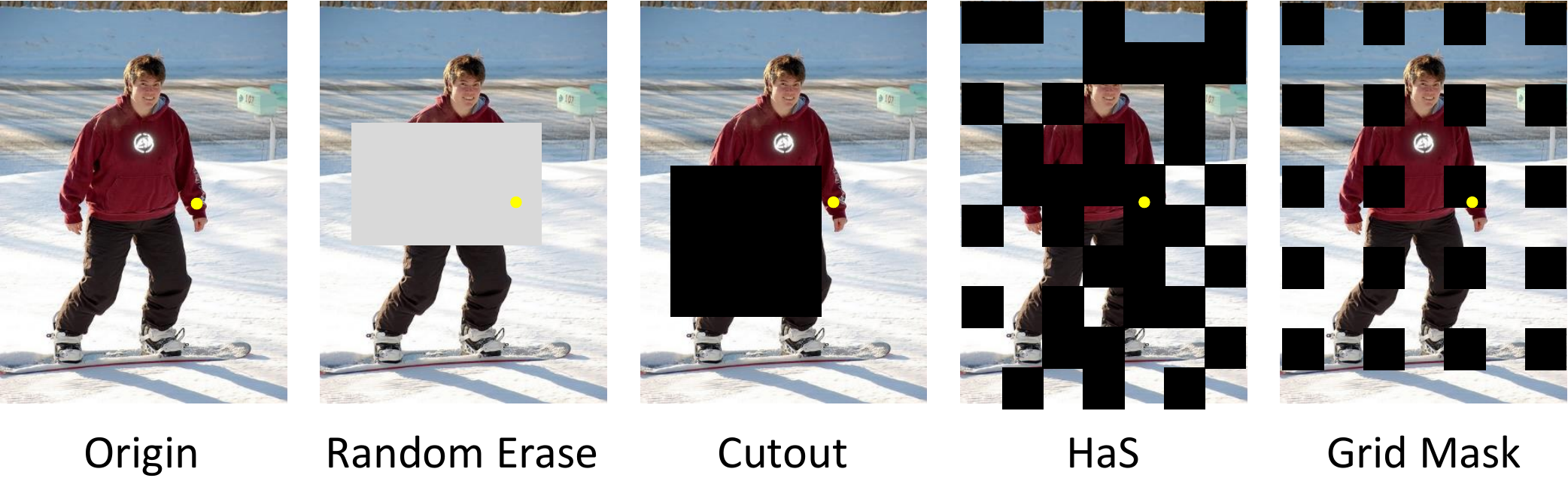}
    \end{center}
   \caption{The illustration of different information dropping methods. Random Erase and Cutout perform single-area information dropping while Hide-and-Seek(HaS) and GridMask perform multi-area information dropping. All of them have a certain probability of dropping the appearance information of keypoints.}
    \label{fig:occlusion}
\end{figure*}

In this paper, we pay attention to the possible \textit{overfitting} problem, which is a conjecture raised in rethinking the relationship between manually labeling methods and the model training supervisions. The base information that human use for keypoint locating is the appearance cues. This inspires the pioneers to use response map, whose center is exactly located at the keypoints, as the supervision in the human pose estimation training process. The response map supervision is intuitive and has been proved effective in most existing works \cite{SBNet,HRNet,MSPN,RSN,DARK,UDP,Higher}. Besides, another cue is the constraints like the keypoint relationship in human pose or the interaction between human and its surrounding environment. Constraint cues enable one locating the keypoints under some challenging situations where appearance cues are absent or not sufficient, such as occlusion or ambiguity between left and right knees. Although the powerful neural networks have potential to learn from the training data, the constraint cue is still too hard for detectors to learn. By contrast, the appearance cue is intuitively easier for acquiring with convolutional neural network. When the appearance cue is always present and there are no penalty on the neglecting of constraint cue, we suspect that the algorithms only with response map supervision have a tendency to overfitting the appearance cue.

Based on the aforementioned analysis, we introduce information dropping methods to verify this conjecture and indirectly force the neural network to focus on the constraint cue learning. Information dropping is a well known method for regularization and has been widely used in many other problems \cite{BTforREID,IDforFR,SENet,IDforDET,REDA,Cutout,HaS}. By dropping information in images, the neural networks can learn discriminative features, resulting in a notable increase of model robustness. Inspired by this, we randomly drop the appearance information of a keypoint and maintain the response map supervision, in purpose of preventing the trained estimators from overfitting the appearance cues and making it pay more attention to the constraints. We call this \texttt{Augmentation by Information Dropping(AID)}.

Although AID is theoretically agreed with the purpose of paying more attention to the constraint cues, the effect is negligible or even negative if we use the common training schedule \cite{MSSAN,TAW}. With the observation of the loss and performance in training process, we find that AID makes the standard learning process much more challenging. Specifically, the appearance information shortage caused by AID challenges the early training process, confusing the network just like letting a child learn quantum mechanics. To address this problem, two customized training schedules are proposed in this paper to provide the prerequisite access for higher performance human pose estimation with AID.

In experiments, we apply AID to the state-of-the-art methods in both top-down and bottom-up paradigms. Without bells and whistles, AID successfully pushes the performance boundary of human pose estimation problem by considerable and stable margin in different input sizes, frameworks, backbones, training and testing sets. On challenging COCO \cite{COCO} human pose estimation \texttt{test} set, AID consistently boosts the performance of various configurations by around 0.6 AP in top-down paradigm and up to 1.5 AP in bottom-up paradigm. On more challenging CrowdPose \cite{Crowdpose} dataset, the improvement is more than 1.5 AP. The experimental results not only verify the potential shortcoming in the state-of-the-arts with response map supervision, but also prove the general effectiveness of AID in performance improvement. Based on the strong results, we recommend AID to be a common configuration for training human pose estimators, which is the same as random flip, random scale and random rotation. In addition with ablation study on information dropping methods and train schedules, we offer some guidelines about performing AID.


The main contributions of this paper can be summarized as follows:
\begin{itemize}
  \item [1.]
  This paper pioneers the diagnosis of the appearance cue overfitting problem in human pose estimation and introduces \texttt{Augmentation by Information Dropping} (AID) to verify and address it.
  \item [2.]
  The inefficiency of AID in previous work is analyzed from the viewpoint of information shortage in early training process. And the proposed customized training schedules in this paper are the prerequisite of performance improvement with AID.
  \item [3.]
  With thorough experiments, we showcase that AID successfully pushes the performance boundary of human pose estimation problem and sets a new state-of-the-art baseline for it. Thus, we hope AID to be a regular configuration for training human pose estimators.
\end{itemize}

\section{Related Work}
\label{sec:RW}
%
\subsection{Human Pose Estimation}
\textbf{Bottom-up methods} detect identity-free keypoints for all the persons at first, and then group them into person instances. Most bottom-up methods focus on the grouping problem. OpenPose \cite{OpenPose} adds another branch to learn pairwise relationships (part affinity fields) between keypoints for grouping. AssociativeEmbedding \cite{AssociativeEmbedding} groups the keypoints just according to the embedding vector which is learned alone with heatmaps. \cite{PIFPAF} proposes to learn the Part Intensity Field, aiming at precisely locating small instance. MultiPoseNet \cite{MultiPoseNet} simultaneously achieves human detection and pose estimation, and proposes PRN to group the keypoints by the bounding box of each people. At the cost of high computation, HigherHRNet \cite{Higher} maintains high-resolution feature maps which effectively improve the precision of the predictions mainly by reducing the systemic error which is stated in UDP \cite{UDP}. \cite{DHGG} replaces the postprocessing grouping with differentiable hierarchical graph grouping to achieve end-to-end learning for multi-person pose estimation.

\textbf{Top-down methods} achieve multi-person pose estimation by the two-stages process: detecting the bounding box of persons by a person detector and perceiving individual keypoint locations within these boxes. The architecture of backbones is the main consideration in this paradigm. CPN \cite{CPN} and MSPN \cite{MSPN} are the leading methods on COCO Keypoint Challenge in 2017 and 2018 respectively, with the main idea of refining keypoint prediction with cascade networks. As a follower, RSN \cite{RSN} designs Res-Steps-Net unit and pose refine machine to learn delicate local representations specific for MSPN network architecture. SimpleBasline \cite{SBNet} proposes a simple but effective paradigm by adding a few deconvolutional layers to enlarge the resolution of output features. HRNet \cite{HRNet} maintains high-resolution representations through the whole architecture, achieving state-of-the-art performance on public dataset. Mask R-CNN \cite{Mask-RCNN} achieves a good balance between performance and inference speed by building an end-to-end framework. PoseFix \cite{Posefix} is designed as a postprocessing module that learns to modify the mistake in existing methods. Analogously, Graph-PCNN \cite{Graph-PCNN} designs an extra refine stage which revises the feature for localization and takes the relationship between keypoints into consideration. Recently, some other works pay attention to the data processing aspect of human pose estimation. DARK \cite{DARK} achieves high precision decoding by designing a distribution-aware method. UDP \cite{UDP} diagnoses the bias data processing in existing methods and form a higher and more reliable baseline for human pose estimation problem, being the strong basic of the champion solution on COCO Keypoint Challenge in 2020.

Some previous works explicitly utilize the constraint cue in neural network construction or postprocessing. By using predefined pose graph and complex neural network architecture, \cite{SCI} designs Cascade Prediction Fusion and Pose Graph Neural Network to exploit underlying contextual information. OpenPose \cite{OpenPose} builds a model that contains two branches to predict keypoint heatmaps and pairwise relationships (part affinity fields) between them. The part affinity fields explicitly learn the constraint cue and are used in grouping process. Whether this constraint cue can promote the response map predicting or not has not been studied. When using the constraint cue, the aforementioned methods only focus on the model of human body, and few jobs consider the interaction between target and environment.

\subsection{Information dropping}
As an effective way for regularization, information dropping has served as a common training strategy for many tasks such as person re-identification \cite{BTforREID}, face recognition \cite{IDforFR}, classification \cite{SENet} and object detection \cite{IDforDET}. Information dropping for deep learning derives from random erasing \cite{REDA} and Cutout \cite{Cutout}. Recently, hide-and-seek (HaS) \cite{HaS} and GridMask \cite{GridMask} put forward two multi-area information dropping strategies respectively, achieving better regularization effect in classification problem.

To the best of our knowledge, augmentation by information dropping has not been a popular paradigm in human pose estimation problem and is absent in most state-of-the-arts. \cite{MSSAN} pioneers keypoint masking training in purpose of imitating different occlusion situation, however the improvement is negligible. \cite{TAW} performs thorough experiments to prove that occlusion augmentation is not necessary for human pose estimation, as it provides no improvement or even degrades the performance under some situations. The previous works use the same training schedule for fair comparison when perform ablation study on augmentation with information dropping or information disturbing. We will analyze this from the viewpoint of information supplying in training process, which results in a conclusion that the best schedules are different in training with or without information dropping augmentation.

\begin{figure*}[t]
	\setlength{\abovecaptionskip}{0.cm}
    \begin{center}
        \includegraphics[width=0.90\hsize]{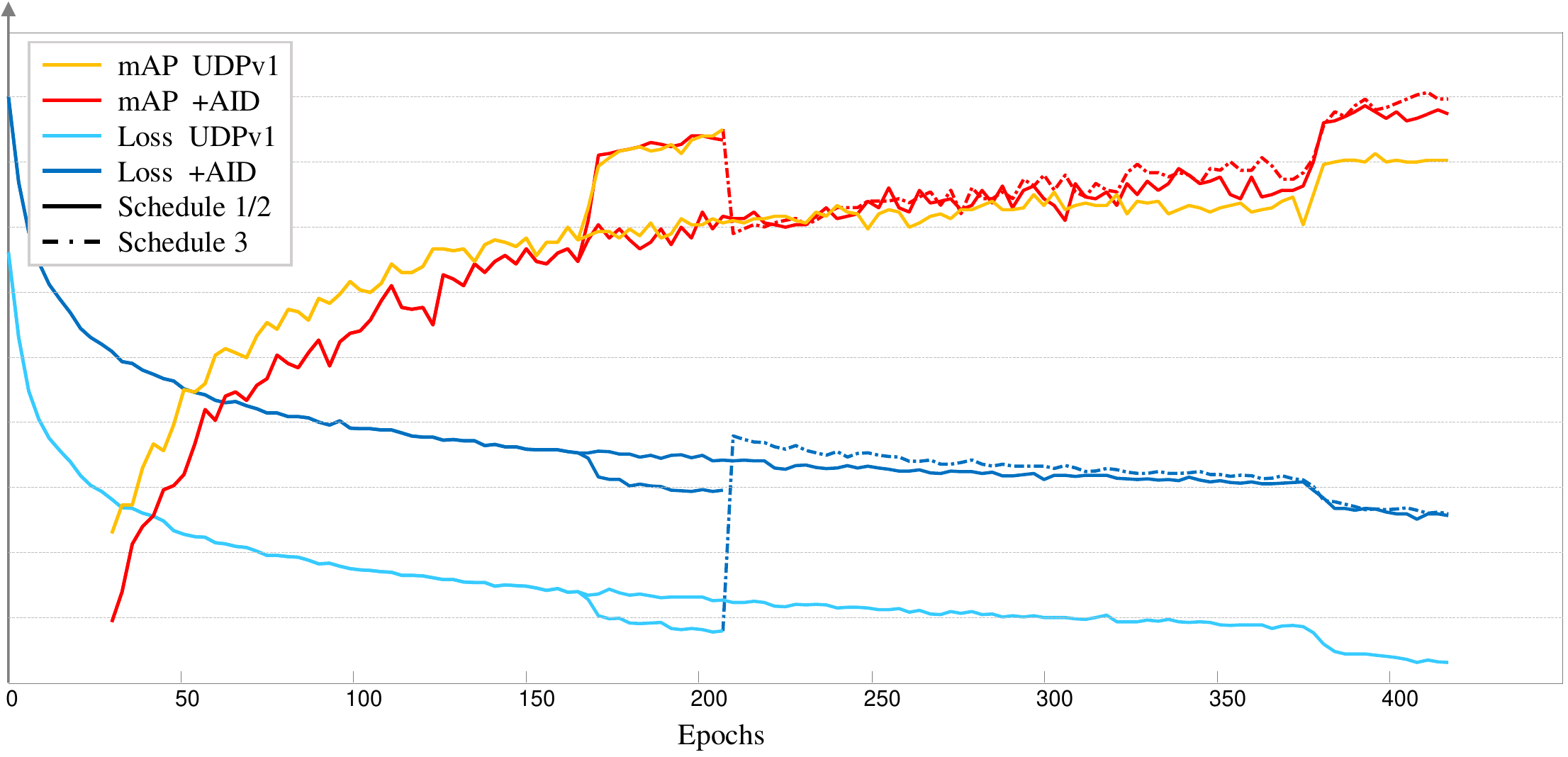}
    \end{center}
   \caption{The training loss and performance of different configurations. Training loss with AID is much higher than that without. Trained with AID, the performance of the models falls behind in early phase but gradually catch up with the baseline(UDPv1) and becomes superior in the later. AID degrades the performance in early stage and postpones the saturation in training process. However it effectively pushes the performance boundary of human pose estimation when the training reaches saturation.  }
    \label{fig:map}

\end{figure*}

\section{Methodology}
\label{sec:MT}
\subsection{AID for Human Pose Estimation}
The idea of information dropping is to randomly drop appearance information of some specific annotated keypoints while maintaining the response map supervision, keeping the training process away from overfitting. As illustrated in Figure \ref{fig:occlusion}, random erase \cite{REDA} and Cutout \cite{Cutout} achieve this by dropping a single continuous area centered at a random position in the image plane. By contrast, HaS \cite{HaS} and GridMask \cite{GridMask} perform multi-area information dropping. HaS firstly splits the image into small patches and then drops some of them under a certain probability. GridMask drops appearance information according to a regular mask constructed with uniformly distributed squares. All aforementioned methods have a certain probability of dropping the appearance information of keypoints, but have different effect on the performance. According to preliminary experiment results, Cutout is the superior methods for top-down paradigm while HaS for bottom-up paradigm. We offer more details about ablation study in Section \ref{sec:ablation}.



\subsection{Training Schedule for AID}
Optimisation schedule is of significance for training high performance pose estimation networks with AID. Empirically, directly applying AID in training process even degrades the performance of pose estimator. By performing ablation study, we observe the variation of loss and performance in training process as illustrated in figure~\ref{fig:map}. The train loss is much higher when training with AID and the performance is lagging in early training process but gradually catch up with in the following. Base on this observation, we argue that the shortage of appearance information caused by AID disturbs the early study and postpones the saturation. Thus training human pose estimators with AID requires a longer schedule.

Here we propose two simple but effective ways to tackle this dilemma. One is to double the training schedule, leaving enough time for the network to conquer the difficulty. And the other is to split the training process into two stages: the training process starts with a common schedule without AID as the previous works, followed by an extra refinement schedule as long as the first one with AID. The main advantage of second approach is that we can reuse the existing well training models in previous works to save computational resource. Empirically, the two training schedules mentioned above have the similar effect and an algorithm can gain a proper promotion from either of them.

\begin{table*}
\footnotesize
\begin{center}
\begin{tabular}{l|l|c|lccccccc}

\hline
Method                  & Backbone      &Input size        &AP  & $\text{AP}^{50}$ & $\text{AP}^{75}$ & $\text{AP}^{\text{M}}$ &$\text{AP}^{\text{L}}$    & AR  &AP-vis         &AP-invis\\
\hline
\multicolumn{11}{c}{Bottom-up methods}\\
\hline
HigherHRNet \cite{Higher}& HRNet-W32     &$512\times512$    &64.4                 & -             & -             & 57.1          &75.6           &-             &-              &-   \\
UDPv1 \cite{UDP}         & HRNet-W32     &$512\times512$    &67.0                 & 86.2          & 72.0          & 60.7          &76.7           &71.6          &71.2           &59.9\\
UDPv1\textbf{+AID}      & HRNet-W32     &$512\times512$    &\textbf{68.4 (+1.4)}  &\textbf{88.1}  &\textbf{74.9}  &\textbf{62.7}  &\textbf{77.1}  &\textbf{73.0} &\textbf{72.6}  &\textbf{60.6}\\
HigherHRNet \cite{Higher}& HigherHRNet-W32&$512\times512$   &67.1                 & 86.2          & 73.0          & 61.5          &76.1           & -            &-              &-   \\
UDPv1 \cite{UDP}         & HigherHRNet-W32&$512\times512$   &67.8                 & 86.2          & 72.9          & 62.2          &76.4           &72.4          &72.2           &59.6\\
UDPv1\textbf{+AID}      & HigherHRNet-W32&$512\times512$   &\textbf{69.0 (+1.2)}  &\textbf{88.0}  &\textbf{74.9}  &\textbf{64.0}  &\textbf{76.9}  &\textbf{73.8} &\textbf{73.2}  &\textbf{60.8}\\
HigherHRNet \cite{Higher}& HigherHRNet-W48&$640\times640$   &69.9                 & 87.2          & 76.1          & 65.4          &76.4           & -            &-              &-   \\
UDPv1 \cite{UDP}         & HigherHRNet-W48&$640\times640$   &69.9                 & 87.3          & 76.2          & 65.9          &76.2           &74.4          &74.1           &60.6\\
UDPv1\textbf{+AID}      & HigherHRNet-W48&$640\times640$   &\textbf{71.0 (+1.1)}  &\textbf{88.2}  &\textbf{77.3}  &\textbf{67.4}  &\textbf{77.1}  &\textbf{75.5} &\textbf{75.2}  &\textbf{62.0}\\
\hline
\multicolumn{11}{c}{Bottom-up methods with multi-scale test as in HigherHRNet \cite{Higher}}\\
\hline
UDPv1 \cite{UDP}         & HRNet-W32     &$512\times512$    &70.4                 & 88.2          & 75.8          & 65.3          &77.6           &74.7          &74.5              &62.3\\
UDPv1 \textbf{+AID}      & HRNet-W32     &$512\times512$    &\textbf{71.1 (+0.7)}  &\textbf{88.9}  &\textbf{77.2}  &\textbf{66.7}  &\textbf{77.8}  &\textbf{75.5} &\textbf{75.4}  &\textbf{62.4}\\
HigherHRNet \cite{Higher}& HigherHRNet-W32&$512\times512$   &69.9                 & 87.1          & 76.0          & 65.3          &77.0           & -            &-              &-   \\
UDPv1 \cite{UDP}         & HigherHRNet-W32&$512\times512$   &70.2                 & 88.1          & 76.2          & 65.4          &77.4           &74.5          &74.6           &61.2\\
UDPv1\textbf{+AID}      & HigherHRNet-W32&$512\times512$   &\textbf{71.3 (+1.1)}  &\textbf{89.0}  &\textbf{77.4}  &\textbf{66.9}  &\textbf{77.7}  &\textbf{75.6} &\textbf{75.7}  &\textbf{61.5}\\
HigherHRNet \cite{Higher}& HigherHRNet-W48&$640\times640$   &72.1                 & 88.4          & 78.2          & 67.8          &78.3           & -            &-              &-   \\
UDPv1 \cite{UDP}         & HigherHRNet-W48&$640\times640$   &71.5                 & 88.3          & 77.3          & 67.9          &77.2           &75.9          &76.1           &61.7\\
UDPv1\textbf{+AID}      & HigherHRNet-W48&$640\times640$   &\textbf{73.0 (+1.5)}  &\textbf{89.2}  &\textbf{79.3}  &\textbf{69.2}  &\textbf{78.6}  &\textbf{77.0} &\textbf{77.2}  &\textbf{63.4}\\
\hline
\multicolumn{11}{c}{Top-down methods}\\
\hline
UDPv1 \cite{UDP}         & ResNet-50     &$256\times192$    &74.6                 & 91.0          & 81.8          & 70.9          &81.1           &80.1          &78.3           &67.4\\
\textbf{+AID}            & ResNet-50     &$256\times192$    &\textbf{75.3 (+0.7)}  &\textbf{91.5}  &\textbf{82.8}  &\textbf{71.7}  &\textbf{81.8}  &\textbf{80.9} &\textbf{79.0}  &\textbf{68.2}\\
UDPv1 \cite{UDP}         & 2xRSN-50      &$256\times192$    &77.7                 & 91.7          & 84.7          & 74.3          &84.2           &83.3          &81.0           &\textbf{70.8}\\
\textbf{+AID}            & 2xRSN-50      &$256\times192$    &\textbf{78.2 (+0.5)}  &\textbf{92.1}  &\textbf{84.7}  &\textbf{74.6}  &\textbf{84.6}  &\textbf{83.4} &\textbf{81.5}  &70.7\\
UDPv1 \cite{UDP}         & HRNet-W32     &$256\times192$    &77.2                 & 91.6          & 84.2          & 73.7          &83.7           &82.5          &80.7           &69.4\\
\textbf{+AID}            & HRNet-W32     &$256\times192$    &\textbf{77.8 (+0.6)}  & \textbf{92.1} & \textbf{84.5} & \textbf{74.1} &\textbf{84.1}  &\textbf{82.8} &\textbf{81.1}  &\textbf{70.3}\\
UDPv1 \cite{UDP}         & HRNet-W48     &$256\times192$    &77.8                 & 92.0          & 84.2          & 74.4          &84.1           &83.0          &81.3           &70.3\\
\textbf{+AID }           & HRNet-W48     &$256\times192$    &\textbf{78.4 (+0.6)}  & \textbf{92.3} & \textbf{84.9} & \textbf{75.1} &\textbf{84.6}  &\textbf{83.4} &\textbf{81.7}  &\textbf{70.8}\\
UDPv1 \cite{UDP}         & HRNet-W32     &$384\times288$    &77.9                 & 91.7          & 83.9          & 74.1          &84.5           &83.1          &81.4           &70.5\\
\textbf{+AID}            & HRNet-W32     &$384\times288$    &\textbf{78.7 (+0.8)}  & \textbf{92.2} & \textbf{85.0} & \textbf{75.0} &\textbf{85.1}  &\textbf{83.6} &\textbf{81.9}  &\textbf{71.4}\\
UDPv1 \cite{UDP}         & HRNet-W48plus &$384\times288$    &78.5                 & 91.9          & 84.9          & 74.6          &85.2           &83.6          &81.7           &71.5\\
\textbf{+AID }           & HRNet-W48plus &$384\times288$    &\textbf{79.1 (+0.6)}  & \textbf{92.2} & \textbf{85.3} & \textbf{75.4} &\textbf{85.7}  &\textbf{84.1} &\textbf{82.3}  &\textbf{71.8}\\
\hline

\hline
\end{tabular}
\end{center}
\caption{Comparisons on COCO \texttt{val} set. AID consistently boosts the performance of the state-of-the-arts by around 0.6 AP in top-down paradigm and up to 1.5 AP in bottom-up paradigm. HRNet-W48plus: A modification of HRNet-W48 with deeper network structure.}
\label{tab:val}
\end{table*}

\begin{table*}
\footnotesize
\begin{center}
\begin{tabular}{l|l|c|lcccccc}

\hline
Method                           & Backbone       &Input size      &AP   & $\text{AP}^{50}$ & $\text{AP}^{75}$ & $\text{AP}^{\text{M}}$ &$\text{AP}^{\text{L}}$ &AR  \\
\hline
\multicolumn{9}{c}{Bottom-up methods}\\
\hline
Hourglass \cite{AssociativeEmbedding}& Hourglass   &$512\times512$    &56.6                  & 81.8              & 61.8         & 49.8          &67.0           &-   \\
PersonLab \cite{PersonLab}         & ResNet-152    &$1401\times1401$  &66.5                  & 88.0              & 72.6         & 62.4          &72.3           &-\\
PifPaf \cite{PIFPAF}               & -             &-                 &66.7                  & -                 & -            &-              &-              &-    \\
HigherHRNet \cite{Higher}          & HRNet-W32     &$512\times512$    &64.1                  & 86.3              & 70.4         & 57.4          &73.9           &-\\
UDPv1 \cite{UDP}                   & HRNet-W32     &$512\times512$    &66.8                  & 88.2              & 73.0         & 61.1          &75.0           &71.5\\
UDPv1\textbf{+AID}                & HRNet-W32     &$512\times512$    &\textbf{67.8 (+1.0)}   &\textbf{89.2}      &\textbf{74.0} &\textbf{62.3}  &\textbf{75.6}  &\textbf{72.5}\\
HigherHRNet \cite{Higher}          &HigherHRNet-W32&$512\times512$    &66.4                  & 87.5              & 72.8         & 61.2          &74.2           &-\\
UDPv1 \cite{UDP}                   &HigherHRNet-W32&$512\times512$    &67.2                  & 88.1              & 73.6         & 62.0          &74.3           &72.0\\
UDPv1\textbf{+AID}                &HigherHRNet-W32&$512\times512$    &\textbf{68.1 (+0.9)}   &\textbf{89.1}      &\textbf{74.7} &\textbf{63.4}  &\textbf{74.9}  &\textbf{73.1}\\
HigherHRNet \cite{Higher}          &HigherHRNet-W48&$640\times640$    &68.4                  & 88.2              & 75.1         & 64.4          &74.2           &-\\
UDPv1\cite{UDP}                   &HigherHRNet-W48&$640\times640$    &68.6                  & 88.2              & 75.5         & 65.0          &74.0           &73.5\\
UDPv1\textbf{+AID}                &HigherHRNet-W48&$640\times640$    &\textbf{70.1(+1.5)}   &\textbf{89.3}      &\textbf{76.8} &\textbf{66.5}  &\textbf{74.9}  &\textbf{74.7}\\
\hline
\multicolumn{9}{c}{Bottom-up methods with multi-scale test as in HigherHRNet \cite{Higher}}\\
\hline

UDPv1 \cite{UDP}                   & HRNet-W32     &$512\times512$    &69.3                  & 89.2              & 76.0         & 64.8          &76.0           &74.1\\
\textbf{+AID}                     & HRNet-W32     &$512\times512$    &\textbf{70.2 (+0.9)}   &\textbf{90.1}      &\textbf{77.1} &\textbf{65.7}  &\textbf{76.4}  &\textbf{74.8}\\
UDPv1 \cite{UDP}                   &HigherHRNet-W32&$512\times512$    &69.1                  & 89.1              & 75.8         & 64.4          &75.5           &73.8\\
\textbf{+AID}                     &HigherHRNet-W32&$512\times512$    &\textbf{69.9 (+0.8)}   &\textbf{89.3}      &\textbf{76.8} &\textbf{65.6}  &\textbf{75.8}  &\textbf{74.6}\\
HigherHRNet \cite{Higher}          &HigherHRNet-W48&$640\times640$    &70.5                  & 89.3              & 77.2         & 66.6          &75.8           &-\\
UDPv1\cite{UDP}                   &HigherHRNet-W48&$640\times640$    &70.5                  & 89.4              & 77.0         & 66.8          &75.4           &75.1\\
\textbf{+AID}                     &HigherHRNet-W48&$640\times640$    &\textbf{71.5(+1.0)}   &\textbf{90.2}      &\textbf{78.0} &\textbf{67.8}  &\textbf{76.2}  &\textbf{76.1}\\
\hline
\multicolumn{9}{c}{Top-down methods}\\
\hline
Mask-RCNN \cite{Mask-RCNN}        & ResNet-50-FPN  &-                &63.1                  & 87.3             & 68.7          & 57.8          &71.4           &-   \\
Integral Pose Regression \cite{IPR}& ResNet-101    &$256\times 256$  &67.8                  & 88.2             & 74.8          & 63.9          &74.0           &-   \\
G-RMI+extra data \cite{G-RMI}     & ResNet-101       &$353\times 257$&68.5                  & 87.1             & 75.5          & 65.8          &73.3           &73.3\\
RMPE \cite{RMPE}             & PyraNet\cite{PyraNet}&$320\times 256$ &72.3                  & 89.2             & 79.1          & 68.0          &78.6           &-   \\
CFN \cite{CFN}                    & -                &-              &72.6                  & 86.1             & 69.7          & 78.3          &64.1           &-   \\
CPN(ensemble) \cite{CPN}          & ResNet-Inception &$384\times 288$&73.0                  & 91.7             & 80.9          & 69.5          &78.1           &79.0\\
CSANet \cite{CSANet}              & ResNet-152    &$384\times288$    &74.5                  & 91.7             & 82.1          & 71.2          &80.2           &80.7\\
MSPN* \cite{MSPN}                 & MSPN          &$384\times288$    &77.1                  & 93.8             & 84.6          & 73.4          &82.3           &82.3\\
HRNet \cite{HRNet}                & HRNet-W32     &$384\times288$    &74.9                  & 92.5             & 82.8          & 71.3          &80.9           &80.1\\
HRNet \cite{HRNet}                & HRNet-W48     &$384\times288$    &75.5                  & 92.5             & 83.3          & 71.9          &81.5           &80.5\\
HRNet* \cite{HRNet}               & HRNet-W48     &$384\times288$    &77.0                  & 92.7             & 84.5          & 73.4          &83.1           &82.0\\
DARK \cite{DARK}                  & HRNet-W48     &$384\times288$    &76.2                  & 92.5             & 83.6          & 72.5          &82.4           &81.1\\
DARK* \cite{DARK}                 & HRNet-W48     &$384\times288$    &77.4                  & 92.6             & 84.6          & 73.6          &83.7           &82.3\\
RSN \cite{RSN}                   & 2xRSN-50       &$256\times192$    &75.5                  & 93.6             & 84.0          & 73.0          &79.6           &81.3\\
PoseFix \cite{Posefix}    & HRNet-W48+ResNet-152    &$384\times288$  &76.7                  & 92.6             & 84.1          & 73.1          &82.6           &81.5\\
Graph-PCNN \cite{Graph-PCNN}      & HRNet-W48     &$384\times288$    &76.8                  & 92.6             & 84.3          & 73.3          &82.7           &81.6\\

\hline
UDPv1 \cite{UDP}                  & ResNet-50     &$256\times192$    &73.1                  & 91.9              & 80.9         & 69.6          &78.9           &79.1\\
\textbf{+AID}                     & ResNet-50     &$256\times192$    &\textbf{73.7 (+0.6)}   &\textbf{92.2}      &\textbf{81.6} &\textbf{70.4}  &\textbf{79.4}  &\textbf{79.7}\\
UDPv1 \cite{UDP}                   & 2xRSN-50      &$256\times192$    &76.0                  & 92.5              & 83.6         & 73.2          &81.4           &82.3\\
\textbf{+AID}                     & 2xRSN-50      &$256\times192$    &\textbf{76.6 (+0.6)}   &\textbf{92.8}      &\textbf{84.3} &\textbf{73.6}  &\textbf{82.1}  &\textbf{82.5}\\
UDPv1 \cite{UDP}                  & HRNet-W32     &$256\times192$    &75.6                  & 92.3             & 83.2          & 72.3          &81.4           &81.4\\
\textbf{+AID}                     & HRNet-W32     &$256\times192$    &\textbf{76.2 (+0.6)}   & \textbf{92.8}    &\textbf{83.8}  & \textbf{72.9} &\textbf{81.8}  &\textbf{81.8}\\
UDPv1 \cite{UDP}                  & HRNet-W32     &$384\times288$    &76.5                  & 92.6             & 83.8          & 72.9          &82.5           &82.1\\
\textbf{+AID}                     & HRNet-W32     &$384\times288$    &\textbf{77.0 (+0.5)}   &\textbf{92.9}     & \textbf{84.5} & \textbf{73.6} &\textbf{82.8}  &\textbf{82.5}\\
UDPv1 \cite{UDP}                  & HRNet-W48     &$256\times192$    &76.1                  & 92.4             & 83.7          & 72.7          &81.9           &81.8\\
\textbf{+AID}                     & HRNet-W48     &$256\times192$    &\textbf{76.7 (+0.6)}   & \textbf{92.9}    & \textbf{84.2} &\textbf{73.4}  &\textbf{82.3}  &\textbf{82.3}\\
UDPv1 \cite{UDP}                  & HRNet-W48plus &$384\times288$    &76.8                  & 92.8             & 84.1          & 73.4          &82.6           &82.4\\
\textbf{+AID}                     & HRNet-W48plus &$384\times288$    &\textbf{77.5 (+0.7)}   & \textbf{92.9}    & \textbf{84.9} & \textbf{74.2} &\textbf{83.4}  &\textbf{83.0}\\
UDPv1* \cite{UDP}                 & HRNet-W48plus &$384\times288$    &78.2                  & 92.8             & 85.4          & 74.8          &84.2           &83.6\\
\textbf{+AID}              & HRNet-W48plus &$384\times288$    &\textbf{78.7 (+0.5)}   & \textbf{93.1}    & \textbf{85.9} & \textbf{75.2} &\textbf{84.5}  &\textbf{84.0}\\

\hline
\end{tabular}
\end{center}
\caption{The improvement of AP on COCO \texttt{test-dev} set when the proposed AID is applied to the state-of-the-art methods. * means extra data is used. HRNet-W48plus: A modification of HRNet-W48 with deeper network structure.}
\label{tab:test-dev}
\end{table*}

\begin{figure*}[htb]
	\setlength{\abovecaptionskip}{0.cm}
    \begin{center}
        \includegraphics[width=0.89\hsize]{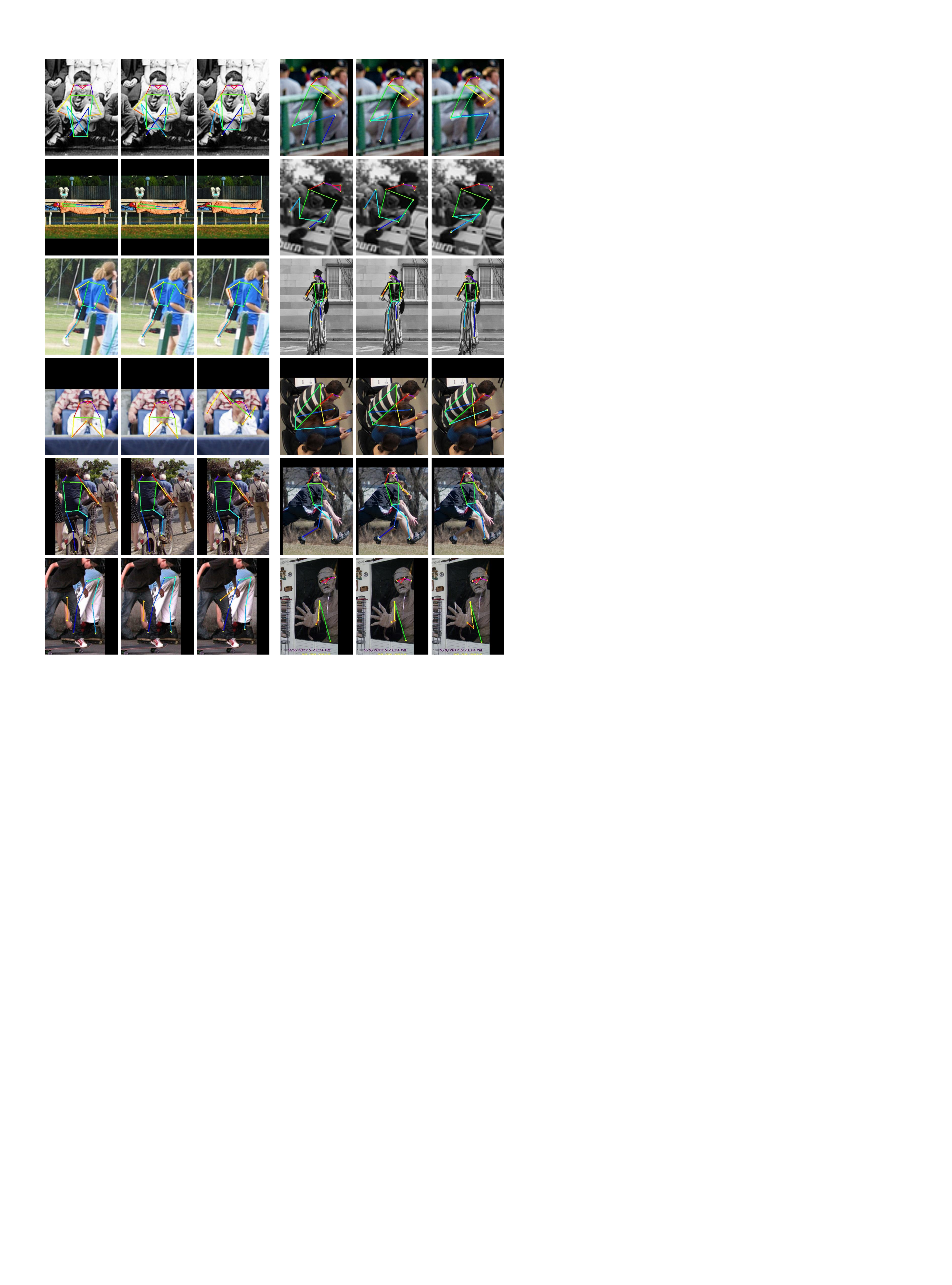}
    \end{center}
   \caption{The visualization of some predicted results under challenging situations. From left to right are ground truths, predicted results with information dropping augmentation and predicted results without information dropping augmentation. }
    \label{fig:illustration}

\end{figure*}

\section{Experiments}
\label{sec:EP}
\subsection{COCO}
\label{sec:coco}
\paragraph{Implementation details.}
Our model is trained on COCO \texttt{train} subset, which is equipped about 57,000 images and 150,000 person instances. We evaluate the trained models on the \texttt{val} set and \texttt{test-dev} set, containing about 5,000 images and 20,000 images, respectively. The AP evaluation metric is reported based on Object Keypoint Similarity (OKS). We use UDPv1 \cite{UDP} as the data-processing guider and set the training configuration strictly following the HRNet-UDPv1 \cite{UDP} for all architectures. State-of-the-art backbones are used in experiments including SimpleBaseline \cite{SBNet}, HRNet \cite{HRNet} and RSN \cite{RSN} for top-down paradigm, HRNet \cite{HRNet} and HigherHRNet \cite{Higher} for bottom-up paradigm. During inference, HTC \cite{HTC} detector is used to detect human instances for top-down paradigm as UDP \cite{UDP}. With multi-scale test, the 80-class and person AP on COCO \texttt{val} set \cite{COCO} are $52.9$ and $65.1$, respectively. We report the performance of single model, and only flipping test strategy is used.

\paragraph{Results on the \texttt{val} set.}
The results of proposed method and state-of-the-arts are listed in Table~\ref{tab:val}. For top-down paradigm, we report the performance improvement when AID is applied to most state-of-the-art architectures including RSN \cite{RSN}, SimpleBaseline \cite{SBNet} and HRNet \cite{HRNet}. The improvement is steady around 0.6 AP among different network architectures. With configurations of HRNet-W32-384$\times$288 and HRNet-W48-256$\times$192, we show that the effect of AID is consistently around +0.6 AP among different network sizes or input sizes. Based on higher baseline with HRNet-W48plus-384$\times$288 configuration, the proposed AID still brings in 0.6 AP improvement. The consistency in performance improvement proves both the widely existing overfitting problem and the universal effectiveness of the proposed augmentation method. For bottom-up paradigm, we take HigherHRNet \cite{Higher} as baseline. AID promotes different configurations by more than 1.1 AP. With multi-scale testing the improvement is up to 1.5 AP.

It is worth noting that AID not only brings steady improvement on the primary metric AP, but also boosts the performance of the algorithms on all other metrics. As AID creates more invisible keypoints and intuitively benefits the perception of them, in addition to study how AID effects the performance on visible and invisible keypoints respectively, we split the original \texttt{val} set into two subsets: \texttt{val-vis} only contains visible keypoints and \texttt{val-invis} only contains invisible keypoints. Based on the predefined metrics in COCO \cite{COCO}, two primary metrics AP-vis and AP-invis are reported in Table~\ref{tab:val} for the two subsets. The experimental results show that, under most configurations, AID not only promotes the performance on the invisible keypoints, but also benefits the perception of the visible keypoints.


\paragraph{Results on the \texttt{test-dev} set.} Table~\ref{tab:test-dev} reports the performance boost of AID on COCO \texttt{test-dev} set. The results show similar improvement compared with \texttt{val} set, indicating the superior generalization property of AID. In addition, we use extra data from AI Challenger \cite{AIchallenger} to verify the effect of AID under the condition of using more training data. With extra training data, the configuration of HRNet-W48plus-384$\times$288-UDPv1 scores 78.2 AP on \texttt{test-dev} set. Although the baseline is unprecedentedly high, the proposed AID still boosts the performance of this configuration by 0.5 AP to 78.7 AP, indicating that more training data effectively improves the performance of trained models but can not tackle the overfitting problem.

\subsection{CrowdPose}
Compared with COCO \cite{COCO}, CrowdPose \cite{Crowdpose} contains more crowded scenes where constraint cues are more required. We take HigherHRNet \cite{Higher} as baseline and maintain their configurations for training and testing. To be specific, the networks are trained on \texttt{train} and \texttt{val} sets with 12k images in total, and the results are reported on \texttt{test} set with 8k images. As listed in Table~\ref{tab:crowdpose}, AID improves HigherHRNet-W32-512$\times$512 configuration by 1.7 AP to 67.3 AP and HigherHRNet-W48-640$\times$640 configuration by 1.5 AP to 68.2 AP. Beside the primary metric, the improvements are also showcased in all the other metrics with considerable margin. What's more, owing to the insufficiency of appearance cue in crowded scenes, the improvements in CrowdPose dataset are more than that in COCO dataset.

\begin{table*}
\footnotesize
\begin{center}
\begin{tabular}{l|l|c|lcccccc}

\hline
Method                           & Backbone      &Input size             &AP                     & $\text{AP}^{50}$ & $\text{AP}^{75}$   & $\text{AP}^{E}$   & $\text{AR}^{M}$       &$\text{AR}^{H}$ \\

\hline
SPPE \cite{Crowdpose}             & ResNet-101    &$320\times240$         &66.0                   & 84.2             & 71.5               & 75.5              &66.3                   &57.4\\
UDPv1 \cite{UDP}                  &HigherHrnet-W32&$512\times512$         &65.6                   & 86.5             & 70.5               & 73.1              &66.2                   &57.5\\
\textbf{+AID}                    &HigherHrnet-W32&$512\times512$         &\textbf{67.3 (+1.7)}    & \textbf{87.3}    &\textbf{72.5}       & \textbf{74.7}     &\textbf{67.9}          &\textbf{59.1}\\
HigherHRNet \cite{Higher}         &HigherHrnet-W48&$640\times640$         &65.9                   & 86.4             & 70.6               & 73.3              &66.5                   &57.9\\
UDPv1 \cite{UDP}                  &HigherHrnet-W48&$640\times640$         &66.7                   & 86.6             & 71.7               & 74.2              &67.3                   &59.1\\
\textbf{+AID}                    &HigherHrnet-W48&$640\times640$         &\textbf{68.2 (+1.5)}    & \textbf{87.3}    &\textbf{73.6}       & \textbf{75.4}     &\textbf{68.6}          &\textbf{60.4}\\
\hline
HigherHRNet* \cite{Higher}        &HigherHrnet-W48&$640\times640$         &67.6                   & 87.4             & 72.6               & 75.8              &68.1                   &58.9\\
UDPv1* \cite{UDP}                  &HigherHrnet-W48&$640\times640$         &68.2                   & 88.0             & 72.9               & 76.6              &68.7                   &59.9\\
\textbf{+AID}*                   &HigherHrnet-W48&$640\times640$         &\textbf{69.7 (+1.5)}    & \textbf{88.4}    &\textbf{74.9}       & \textbf{77.9}     &\textbf{70.3}          &\textbf{61.4}\\
\hline
\end{tabular}
\end{center}
\caption{The improvement of AP on CrowdPose \texttt{test} set when AID is applied. * means multi-scale testing is used.}
\label{tab:crowdpose}
\end{table*}

\subsection{Ablation Study}
\label{sec:ablation}

\subsubsection{Information Dropping Methods}
We perform information dropping method Cutout \cite{Cutout}, HaS \cite{HaS} and GridMask \cite{GridMask} with the implementations from their official projects. As the hyper-parameters is vital for the effectiveness, we adjust them with principle of keeping the initial loss close with each other to provide similar regularization effect. And this offer us a quick access to searching the superior hyper-parameters for each of them. With limited hyper-parameter searching experiments, we report the performance of the best configuration in Table~\ref{tab:diffid}.

We summarize two discoveries here: (a) Performance of different information dropping methods is various in bottom-up paradigm while is close in top-down paradigm. (b) The best method for information dropping is different in different paradigms: HaS for bottom-up with large superiority, while Cutout for top-down with small superiority.

\begin{table}[h]
\footnotesize
\begin{center}
\begin{tabular}{l|c|c|c|c}

\hline
Method     & Baseline    &Cutout        &HaS            &GridMask        \\
\hline
Bottom-up  & 67.8        &68.1          &\textbf{69.0}  &68.2           \\
Top-down   & 77.2        &\textbf{77.8} &77.7           &77.6           \\
\hline
\end{tabular}
\end{center}
\caption{Comparison of different information dropping methods. Results are primary metric AP conducted on COCO \texttt{val} set with configuration HRNet-W32-256$\times$192 for top-down paradigm and HigherHRNet-W32-512$\times$512 for bottom-up paradigm.}
\label{tab:diffid}
\end{table}
\vspace{-0.3cm}

\subsubsection{Training Schedule}
In this subsection, we use top-down paradigm with HRNet-W32-256$\times$192 configurations and ground-truth human boxes. To explore the effect of training schedule on AID, we firstly design three different training schedules:
\begin{itemize}
  \item [S1.]
  Normal training schedule from HRNet \cite{HRNet} with a base learning rate of 1e-3 and is dropped to 1e-4 and 1e-5 at the 170th and 200th epochs, repectively. The training process is terminated within 210 epochs.
  \item [S2.]
  Double the length of the schedule S1. The learning rate is dropped at 380th and 410th epochs, respectively. The training process is terminated within 420 epochs.
  \item [S3.]
  Repeat the schedule S1 twice with different configurations, i.e. first 210 epochs are trained without AID, and applying AID on the subsequent.
\end{itemize}

\begin{table}[h]
\footnotesize
\begin{center}
\begin{tabular}{l|c|c|c|c|c}

\hline
ID                              & E1    &E2   &E3 &E4 &E5        \\
\hline
schedule                        & S1    &S1   &S2 &S2 &S3\\
\hline
AID                             & OFF   &ON   &OFF&ON &OFF/ON  \\
\hline
\end{tabular}
\end{center}
\caption{Configurations of different training schedules for ablation study. }
\label{tab:config}
\end{table}

Based on the pre-defined training schedules, five experimental configurations are constructed as listed in Table~\ref{tab:config}. The performance and the corresponding training loss on COCO \texttt{val} AP metric is illustrated in Figure~\ref{fig:map}. Compared E1 with E2, where the same standard training schedule S1 is used and the variable is whether or not to use AID, the training loss is higher when training with AID. The performance of E2 is poorer than E1 in the early training process, indicating that appearance information is vital in early training process and AID would disturb the study the of appearance feature. However, E1 and E2 have similar performance at the end of this training schedule. This means that, AID will not provide positive effect with the standard training schedule. Compare E3 with E4, where a longer schedule is adopted. The performance of E4 with AID starts surpassing E3 at around 250 epoch. And this superiority gradually grows in the subsequent training process. Compare E3 with E1 and E4 with E2, a longer schedule enables the algorithms learning more useful information when AID is used, but makes the algorithms overfitting the training data when AID is absent. Compared E4 with E5, Schedule2 and Schedule3 offer similar improvements, which means that we can reuse the pre-trained models from the previous works and boost their performance by applying another fine tuning process with AID.

\subsubsection{Qualitative Comparison}
To qualitatively showcase the efficiency of the proposed method, Figure~\ref{fig:illustration} visualizes some detection results under challenging situations, where the appearance cue is not sufficient and constraint cue is necessary for keypoint location. The results produced by models trained with AID are more reasonable and precise than that without AID.

\section{Conclusion and Future Work}
\label{sec:CC}
In this paper, we expose and remedy the possible overfitting problem in human pose estimation by proposing Augmentation by Information Dropping. In response to bias in existing jobs and from the perspective of information supplying, we offer a reasonable explanation for the invalid of AID with standard training schedule and propose customized training schedule to effectively exploit the potential of the proposed information dropping augmentation. As a result, AID offers fundamental breakthrough in robust human pose estimation and consistently boosts the performance of state-of-the-arts by a considerable margin. Future works will focus on proper neural network architecture for constraint cue learning and more efficient training schedule or more effective information dropping formats for AID.

{\small
\bibliographystyle{ieee_fullname}
\bibliography{egbib}

\begin{thebibliography}{10}\itemsep=-1pt

\bibitem{PS}
Mykhaylo Andriluka, Stefan Roth, and Bernt Schiele.
\newblock Pictorial structures revisited: People detection and articulated pose
  estimation.
\newblock In {\em CVPR}, 2009.

\bibitem{RSN}
Yuanhao Cai, Zhicheng Wang, Zhengxiong Luo, Binyi Yin, Angang Du, Haoqian Wang,
  Xinyu Zhou, Erjin Zhou, Xiangyu Zhang, and Jian Sun.
\newblock Learning delicate local representations for multi-person pose
  estimation.
\newblock In {\em ECCV}, 2020.

\bibitem{OpenPose}
Zhe Cao, Tomas Simon, Shih-En Wei, and Yaser Sheikh.
\newblock Realtime multi-person 2d pose estimation using part affinity fields.
\newblock In {\em CVPR}, 2017.

\bibitem{carreira2017quo}
Joao Carreira and Andrew Zisserman.
\newblock Quo vadis, action recognition a new model and the kinetics dataset.
\newblock In {\em CVPR}, 2017.

\bibitem{HTC}
Kai Chen, Jiangmiao Pang, Jiaqi Wang, Yu Xiong, Xiaoxiao Li, Shuyang Sun,
  Wansen Feng, Ziwei Liu, Jianping Shi, Wanli Ouyang, et~al.
\newblock Hybrid task cascade for instance segmentation.
\newblock In {\em CVPR}, 2019.

\bibitem{GridMask}
Pengguang Chen.
\newblock Gridmask data augmentation.
\newblock {\em arXiv preprint arXiv:2001.04086}, 2020.

\bibitem{CPN}
Yilun Chen, Zhicheng Wang, Yuxiang Peng, Zhiqiang Zhang, Gang Yu, and Jian Sun.
\newblock Cascaded pyramid network for multi-person pose estimation.
\newblock In {\em CVPR}, 2018.

\bibitem{Higher}
Bowen Cheng, Bin Xiao, Jingdong Wang, Honghui Shi, Thomas~S. Huang, and Lei
  Zhang.
\newblock Higherhrnet: Scale-aware representation learning for bottom-up human
  pose estimation.
\newblock In {\em CVPR}, 2020.

\bibitem{IDforDET}
Cheng Chi, Shifeng Zhang, Junliang Xing, Zhen Lei, Stan~Z Li, Xudong Zou,
  et~al.
\newblock Pedhunter: Occlusion robust pedestrian detector in crowded scenes.
\newblock In {\em AAAI}, 2020.

\bibitem{Cutout}
Terrance DeVries and Graham~W Taylor.
\newblock Improved regularization of convolutional neural networks with cutout.
\newblock {\em arXiv preprint arXiv:1708.04552}, 2017.

\bibitem{DHGG}
Xie Enze, Wang Wenhai, Qian Chen, Ouyang Wanli, and Luo Ping.
\newblock Differentiable hierarchical graph grouping for multi-person pose
  estimation.
\newblock In {\em ECCV}, 2020.

\bibitem{RMPE}
Hao-Shu Fang, Shuqin Xie, Yu-Wing Tai, and Cewu Lu.
\newblock Rmpe: Regional multi-person pose estimation.
\newblock In {\em ICCV}, 2017.

\bibitem{DPM}
Pedro~F Felzenszwalb, Ross~B Girshick, David McAllester, and Deva Ramanan.
\newblock Object detection with discriminatively trained part-based models.
\newblock {\em IEEE Transactions on Pattern Analysis and Machine Intelligence},
  32(9):1627--1645, 2010.

\bibitem{Mask-RCNN}
Kaiming He, Georgia Gkioxari, Piotr Dollar, and Ross Girshick.
\newblock Mask r-cnn.
\newblock In {\em ICCV}, 2017.

\bibitem{SENet}
Jie Hu, Li Shen, and Gang Sun.
\newblock Squeeze-and-excitation networks.
\newblock In {\em CVPR}, 2018.

\bibitem{UDP}
Junjie Huang, Zheng Zhu, Feng Guo, and Guan Huang.
\newblock The devil is in the details: Delving into unbiased data processing
  for human pose estimation.
\newblock In {\em CVPR}, 2020.

\bibitem{CFN}
Shaoli Huang, Mingming Gong, and Dacheng Tao.
\newblock A coarse-fine network for keypoint localization.
\newblock In {\em ICCV}, 2017.

\bibitem{DeeperCut}
Eldar Insafutdinov, Leonid Pishchulin, Bjoern Andres, Mykhaylo Andriluka, and
  Bernt Schiele.
\newblock Deepercut: A deeper, stronger, and faster multi-person pose
  estimation model.
\newblock In {\em ECCV}, 2016.

\bibitem{Graph-PCNN}
Wang Jian, Long Xiang, Gao Yuan, Ding Errui, and Shilei Wen.
\newblock Graph-pcnn: Two stage human pose estimation with graph pose
  refinement.
\newblock In {\em ECCV}, 2020.

\bibitem{MSSAN}
Lipeng Ke, Ming-Ching Chang, Honggang Qi, and Siwei Lyu.
\newblock Multi-scale structure-aware network for human pose estimation.
\newblock In {\em ECCV}, 2018.

\bibitem{MultiPoseNet}
Muhammed Kocabas, Salih Karagoz, and Emre Akbas.
\newblock Multiposenet: Fast multi-person pose estimation using pose residual
  network.
\newblock In {\em ECCV}, 2018.

\bibitem{PIFPAF}
Sven Kreiss, Lorenzo Bertoni, and Alexandre Alahi.
\newblock Pifpaf: Composite fields for human pose estimation.
\newblock In {\em CVPR}, 2019.

\bibitem{Crowdpose}
Jiefeng Li, Can Wang, Hao Zhu, Yihuan Mao, Hao-Shu Fang, and Cewu Lu.
\newblock Crowdpose: Efficient crowded scenes pose estimation and a new
  benchmark.
\newblock In {\em CVPR}, 2019.

\bibitem{li2019state}
Peng Li, Jiabin Zhang, Zheng Zhu, Yanwei Li, Lu Jiang, and Guan Huang.
\newblock State-aware re-identification feature for multi-target multi-camera
  tracking.
\newblock In {\em CVPR Workshops}, 2019.

\bibitem{MSPN}
Wenbo Li, Zhicheng Wang, Binyi Yin, Qixiang Peng, Yuming Du, Tianzi Xiao, Gang
  Yu, Hongtao Lu, Yichen Wei, and Jian Sun.
\newblock Rethinking on multi-stage networks for human pose estimation.
\newblock {\em arXiv preprint arXiv:1901.00148}, 2019.

\bibitem{COCO}
Tsung-Yi Lin, Michael Maire, Serge Belongie, James Hays, Pietro Perona, Deva
  Ramanan, Piotr Doll{\'a}r, and C~Lawrence Zitnick.
\newblock Microsoft coco: Common objects in context.
\newblock In {\em ECCV}, 2014.

\bibitem{BTforREID}
Hao Luo, Youzhi Gu, Xingyu Liao, Shenqi Lai, and Wei Jiang.
\newblock Bag of tricks and a strong baseline for deep person
  re-identification.
\newblock In {\em CVPR Workshops}, 2019.

\bibitem{Posefix}
Gyeongsik Moon, Ju~Yong Chang, and Kyoung~Mu Lee.
\newblock Posefix: Model-agnostic general human pose refinement network.
\newblock In {\em CVPR}, 2019.

\bibitem{AssociativeEmbedding}
Alejandro Newell, Zhiao Huang, and Jia Deng.
\newblock Associative embedding: End-to-end learning for joint detection and
  grouping.
\newblock In {\em Advances in Neural Information Processing Systems}, 2017.

\bibitem{Hourglass}
Alejandro Newell, Kaiyu Yang, and Jia Deng.
\newblock Stacked hourglass networks for human pose estimation.
\newblock In {\em ECCV}, 2016.

\bibitem{PersonLab}
George Papandreou, Tyler Zhu, Liang-Chieh Chen, Spyros Gidaris, Jonathan
  Tompson, and Kevin Murphy.
\newblock Personlab: Person pose estimation and instance segmentation with a
  bottom-up, part-based, geometric embedding model.
\newblock In {\em ECCV}, 2018.

\bibitem{G-RMI}
George Papandreou, Tyler Zhu, Nori Kanazawa, Alexander Toshev, Jonathan
  Tompson, Chris Bregler, and Kevin Murphy.
\newblock Towards accurate multi-person pose estimation in the wild.
\newblock In {\em CVPR}, 2017.

\bibitem{DeepCut}
Leonid Pishchulin, Eldar Insafutdinov, Siyu Tang, Bjoern Andres, Mykhaylo
  Andriluka, Peter~V Gehler, and Bernt Schiele.
\newblock Deepcut: Joint subset partition and labeling for multi person pose
  estimation.
\newblock In {\em CVPR}, 2016.

\bibitem{TAW}
Rafal Pytel, Osman~Semih Kayhan, and Jan~C van Gemert.
\newblock Tilting at windmills: Data augmentation for deep pose estimation does
  not help with occlusions.
\newblock {\em arXiv preprint arXiv:2010.10451}, 2020.

\bibitem{IDforFR}
Yichun Shi, Xiang Yu, Kihyuk Sohn, Manmohan Chandraker, and Anil~K Jain.
\newblock Towards universal representation learning for deep face recognition.
\newblock In {\em CVPR}, 2020.

\bibitem{HaS}
Krishna~Kumar Singh, Hao Yu, Aron Sarmasi, Gautam Pradeep, and Yong~Jae Lee.
\newblock Hide-and-seek: A data augmentation technique for weakly-supervised
  localization and beyond.
\newblock {\em arXiv preprint arXiv:1811.02545}, 2018.

\bibitem{HRNet}
Ke Sun, Bin Xiao, Dong Liu, and Jingdong Wang.
\newblock Deep high-resolution representation learning for human pose
  estimation.
\newblock In {\em CVPR}, 2019.

\bibitem{IPR}
Xiao Sun, Bin Xiao, Fangyin Wei, Shuang Liang, and Yichen Wei.
\newblock Integral human pose regression.
\newblock In {\em ECCV}, 2018.

\bibitem{tompson2014joint}
Jonathan~J Tompson, Arjun Jain, Yann LeCun, and Christoph Bregler.
\newblock Joint training of a convolutional network and a graphical model for
  human pose estimation.
\newblock In {\em Advances in Neural Information Processing Systems}, pages
  1799--1807, 2014.

\bibitem{DeepPose}
Alexander Toshev and Christian Szegedy.
\newblock Deeppose: Human pose estimation via deep neural networks.
\newblock In {\em CVPR}, 2014.

\bibitem{CPM}
Shih-En Wei, Varun Ramakrishna, Takeo Kanade, and Yaser Sheikh.
\newblock Convolutional pose machines.
\newblock In {\em CVPR}, 2016.

\bibitem{AIchallenger}
Jiahong Wu, He Zheng, Bo Zhao, Yixin Li, Baoming Yan, Rui Liang, Wenjia Wang,
  Shipei Zhou, Guosen Lin, Yanwei Fu, et~al.
\newblock Ai challenger: A large-scale dataset for going deeper in image
  understanding.
\newblock {\em arXiv preprint arXiv:1711.06475}, 2017.

\bibitem{SBNet}
Bin Xiao, Haiping Wu, and Yichen Wei.
\newblock Simple baselines for human pose estimation and tracking.
\newblock In {\em ECCV}, 2018.

\bibitem{fppose}
Wei Yang, Shuang Li, Wanli Ouyang, Hongsheng Li, and Xiaogang Wang.
\newblock Learning feature pyramids for human pose estimation.
\newblock In {\em ICCV}, 2017.

\bibitem{PyraNet}
Wei Yang, Shuang Li, Wanli Ouyang, Hongsheng Li, and Xiaogang Wang.
\newblock Learning feature pyramids for human pose estimation.
\newblock In {\em ICCV}, 2017.

\bibitem{CSANet}
Dongdong Yu, Kai Su, Xin Geng, and Changhu Wang.
\newblock A context-and-spatial aware network for multi-person pose estimation.
\newblock {\em arXiv preprint arXiv:1905.05355}, 2019.

\bibitem{DARK}
Feng Zhang, Xiatian Zhu, Hanbin Dai, Mao Ye, and Ce Zhu.
\newblock Distribution-aware coordinate representation for human pose
  estimation.
\newblock In {\em CVPR}, 2020.

\bibitem{SCI}
Hong Zhang, Hao Ouyang, Shu Liu, Xiaojuan Qi, Xiaoyong Shen, Ruigang Yang, and
  Jiaya Jia.
\newblock Human pose estimation with spatial contextual information.
\newblock {\em arXiv preprint arXiv:1901.01760}, 2019.

\bibitem{REDA}
Zhun Zhong, Liang Zheng, Guoliang Kang, Shaozi Li, and Yi Yang.
\newblock Random erasing data augmentation.
\newblock In {\em AAAI}, 2020.

\bibitem{zhu2019convolutional}
Jiagang Zhu, Wei Zou, Zheng Zhu, and Yiming Hu.
\newblock Convolutional relation network for skeleton-based action recognition.
\newblock {\em Neurocomputing}, 370:109--117, 2019.

\bibitem{zhu2019action}
Jiagang Zhu, Wei Zou, Zheng Zhu, Liang Xu, and Guan Huang.
\newblock Action machine: Toward person-centric action recognition in videos.
\newblock {\em IEEE Signal Processing Letters}, 26(11):1633--1637, 2019.

\end{thebibliography}
}

\end{document}